\documentclass[journal]{IEEEtran}
\ifCLASSINFOpdf
\else
\fi
\usepackage{subfig}
\usepackage{graphicx}
\usepackage{amsmath,amssymb} 
\usepackage{color}
\usepackage[table]{xcolor}
\usepackage[pagebackref=true,breaklinks=true,letterpaper=true,colorlinks,bookmarks=false]{hyperref}
\newcommand{\etal}{\textit{et al.}}
\DeclareMathOperator*{\argmax}{arg\,max}

\newcommand{\baselinetab}{
  \begin{tabular}{l c c c }
  \hline\noalign{\smallskip}
  Method & Precision & Recall & F-score \\
  \noalign{\smallskip}
  \hline
  \noalign{\smallskip}
  Gomez and Karatzas (2013)~\cite{Gomez2013,kumar2013multi}& 0.64 & 0.58 & 0.61 \\
  Gomez and Karatzas (2013) MSER++& 0.50 & 0.71 & 0.59 \\
  MSER $\mid$ $w_{I}$ & 0.69 & 0.58 & 0.63 \\
  MSER $\mid$ $w_{opt}$ & 0.69 & 0.62 & 0.65 \\
  MSER $\mid$ $w_{I}$ $\mid$ \emph{stopping rule} & 0.76 & 0.60 & 0.67 \\
  MSER $\mid$ $w_{opt}$ $\mid$ \emph{stopping rule} & \textbf{0.77} & 0.62 & 0.69 \\
  \cellcolor{blue!25}MSER++ $\mid$ $w_{opt}$  $\mid$ \emph{stopping rule}& \cellcolor{blue!25}0.75 & \cellcolor{blue!25}0.71 & \cellcolor{blue!25}\textbf{0.73} \\
  MSER++ $\mid$ $w_{opt}, w_{opt_2},w_{opt_3},w_{opt_4}$  $\mid$ \emph{stopping rule}& 0.67 & 0.73 & 0.70 \\
  MSER++ $\mid$ $w_{opt}, w_{opt_2},$\dots$,w_{opt_7}$  $\mid$ \emph{stopping rule}& 0.56 & \textbf{0.74} & 0.64  \\
  \hline
  \label{tab:baselinetab}
  \end{tabular}
}

\begin{document}
%
\title{A Fast Hierarchical Method for Multi-script and Arbitrary Oriented Scene Text Extraction}
%
%
%

\author{Lluis~Gomez,
        and~Dimosthenis~Karatzas,~\IEEEmembership{Member,~IEEE}
\thanks{L. Gomez and D. Karatzas are with the Computer Vision Center, Universitat Autonoma de Barcelona. Edifici O, Campus UAB, 08193 Bellaterra (Cerdanyola) Barcelona, Spain E-mail: {lgomez,dimos}@cvc.uab.cat}}

%
%


\markboth{MANUSCRIPT PREPRINT, July~2014}%
{Gomez and Karatzas: A Fast Hierarchical Method for Multi-script and Arbitrary Oriented Scene Text Extraction}

%



\maketitle

\begin{abstract}
Typography and layout lead to the hierarchical organisation of text in words, text lines, paragraphs. This inherent structure is a key property of text in any script and language, which has nonetheless been minimally leveraged by existing text detection methods.
This paper addresses the problem of text segmentation in natural scenes from a hierarchical perspective.
Contrary to existing methods, we make explicit use of text structure, aiming directly to the detection of region groupings corresponding to text within a hierarchy produced by an agglomerative similarity clustering process over individual regions. We propose an optimal way to construct such an hierarchy introducing a feature space designed to produce text group hypotheses with high recall and a novel stopping rule combining a discriminative classifier and a probabilistic measure of group meaningfulness based in perceptual organization.
Results obtained over four standard datasets, covering text in variable orientations and different languages, demonstrate that our algorithm, while being trained in a single mixed dataset, outperforms state of the art methods in unconstrained scenarios.
\end{abstract}


%
\IEEEpeerreviewmaketitle

\section{Introduction}
%
%
%
%
\IEEEPARstart{T}{he} automated understanding of textual information in natural scene images is receiving increasing attention from computer vision researchers over the last decade. Text localization, extraction and recognition methods have evolved significantly and their accuracy has increased drastically in recent years~\cite{karatzas2013icdar}. However, the problem is far from being considered solved: note that the winner methods in the last ICDAR competition achieve only 66\% and 74\% recall in the tasks of text localization and text segmentation respectively. The main difficulties of the problem stem from the extremely high variability of scene text in terms of scale, rotation, location, physical appearance, and typeface design. Moreover, although standard benchmark datasets have traditionally focussed on horizontally-aligned English text, new datasets have recently appeared covering much more unconstrained scenarios including multi-script and arbitrary oriented text~\cite{yao2012detecting,kumar2013multi}. 

\begin{figure}[t]
\centering
\includegraphics[width=0.99\linewidth]{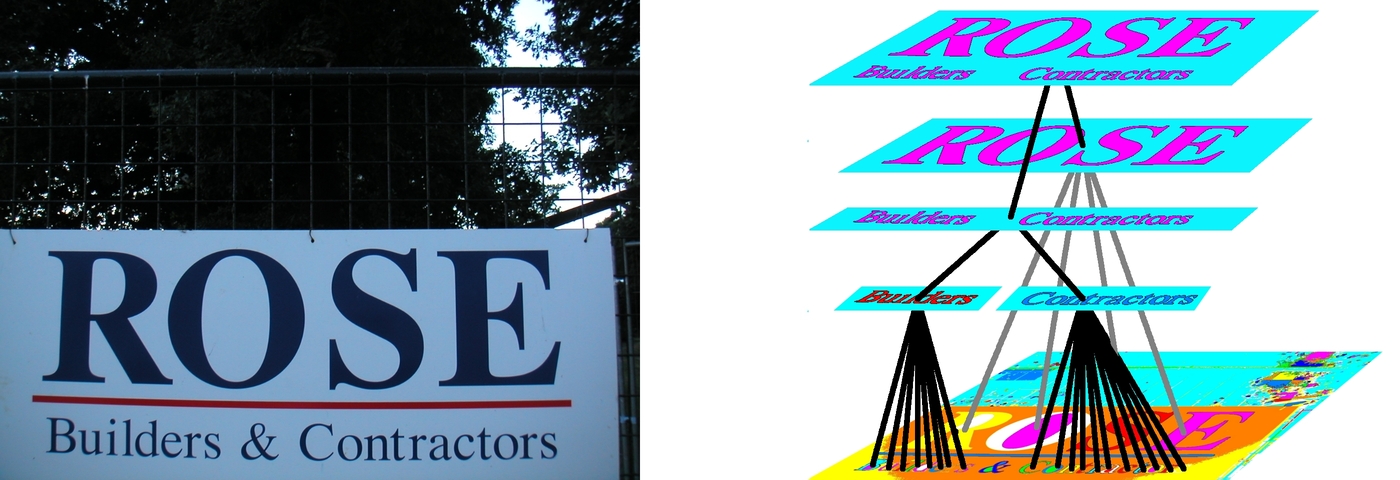}
\caption{A natural scene image and a hierarchical representation of its text. Atomic objects (characters) extracted in the bottom layer are agglomerated into text groupings at different levels of the hierarchy.}
\label{fig:text_hierarchy}
\end{figure}

\begin{figure}[b]
\centering
\includegraphics[width=0.99\linewidth]{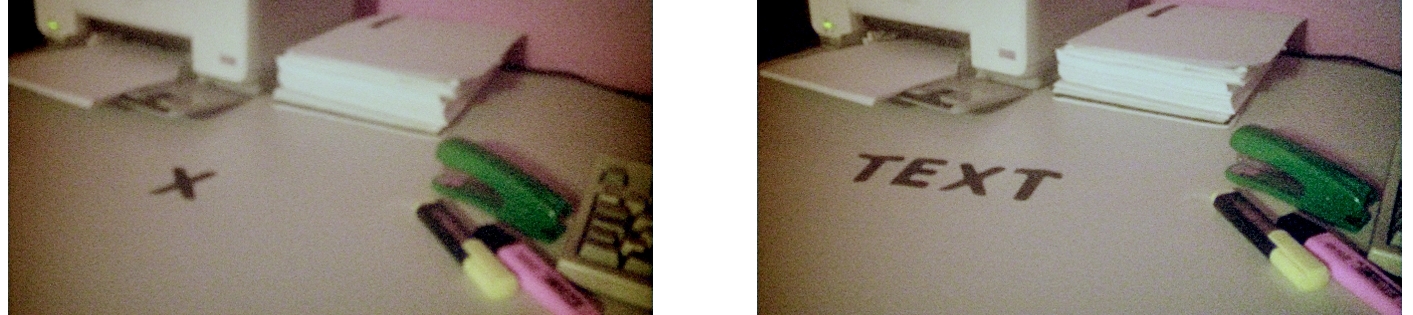}
\caption{Individual text-parts are less distinguishable when viewed separately, but become structurally relevant and easily identifiable when perceived as a group.}
\label{fig:atomic_vs_group}
\end{figure}

Hierarchical organisation is an essential feature of text. Induced by typography and layout the hierarchical arrangement of text strokes leads to the structural formation of text component groupings at different levels (e.g. words, text lines, paragraphs, etc.), see Figure~\ref{fig:text_hierarchy}. This hierarchical property applies independently of the script, language, or style of the glyphs, thus it allows us to pose the problem of text detection in natural scenes in a holistic framework rather than as the classification of individual patches or regions as text or non-text. In fact, as Figure ~\ref{fig:atomic_vs_group} shows, when text-parts are viewed independently out of context they lose their distinguishable text traits, although they become structurally relevant and easily identifiable when observed as a group.

Most existing scene text extraction methods include a grouping step, but more often than not this is done as a heuristic post-processing of regions (e.g. connected components or superpixels) previously classified as text in order to create word or text-line bounding boxes, and it is not part of the core text extraction process~\cite{Epshtein2010,Chen2011}.
Contrary to existing methods, this paper addresses the problem of text segmentation in natural scenes from a hierarchical perspective tackling directly the problem of the detection of groups of text regions, instead of individual regions. The method is driven by an agglomerative clustering process exploiting the strong similarities expected between text components in such groups:
irrespective on the script or language, text is formed by aligned and equally separated glyphs with noticeable contrast to their background, with constant stroke width (thickness), similar color and sizes.

The main contributions of this paper are the following. 
First, we learn an optimal feature space that encodes the similarities between text components thus allowing the Single Linkage Clustering algorithm to generate text group hypotheses with high recall, independently of their orientations and scripts. Second, we couple the hierarchical clustering algorithm with novel discriminative and probabilistic stopping rules, that allow the efficient detection of text groups in a single grouping step. Third, we propose a new set of features for text group classification, that can be efficiently calculated in an incremental way, able to capture the inherent structural properties of text related to the arrangement of text parts and the intra-group similarity.
 
Note that the proposed method is distinctly different from other grouping based state of the art approaches~\cite{Yin2013} in that the text group classification is not an a-posteriori step, but an inherent part of the hierarchical clustering process.
Our findings are positioned in line with recent advances in object recognition~\cite{uijlings2013selective} where bottom-up grouping of an initial segmentation is used to generate object location hypotheses, producing a substantially reduced search space in comparison to the traditional slidding window approaches. We make use of incrementally computable group descriptors in order to make possible the direct evaluation of group hypotheses generated by the clustering algorithm without affecting the overall time complexity of the method. Experiments demonstrate that our algorithm outperforms the state of the art in the MSRRC, MSRA-TD500 and KAIST datasets of multi-script and arbitrary oriented text~\cite{kumar2013multi,yao2012detecting,Lee2010}. 
It is important to notice that our method produces state of the art results in four different datasets with a single (mixed) training set, i.e. it can be seen as a general purpose robust method applicable in many different scenarios. This is afforded by the relatively high-level modelling of text as a group of individual elements, a model which is valid for practically every writing system.

\section{Related Work}

Scene text detection methods can be categorized into texture-based and region-based approaches. Texture-based methods usually work by performing a sliding window search over the image and extracting certain texture features in order to classify each possible patch as text or non-text. Coates \etal~\cite{Coates2011}, and in a different flavour Wang \etal~\cite{Wang2012} and Netzer \etal~\cite{Netzer2011}, propose the use of unsupervised feature learning to generate the features for text versus non-text classification. Wang \etal~\cite{Wang2011}, extending their previous work~\cite{Wang2010}, have built an end-to-end scene text recognition system based on a sliding window character classifier using Random Ferns, with features originating from a HOG descriptor. Mishra \etal~\cite{Mishra2012} propose a closely related end-to-end method based on HOG features and a SVM classifier. Texture based methods yield good text localisation results, although they do not directly address the issue of text segmentation (separation of text from background). Their main drawback compared to region based methods is their lower time performance, as sliding window approaches are confronted with a huge search space in such an unconstrained (i.e. variable scale, rotation, aspect-ratio) task. Moreover, these methods are usually limited to the detection of a single language and orientation for which they have been trained on, therefore they are not directly applicable to the multi-script and arbitrary oriented text scenario.

Region-based methods, on the other hand, are based on a typical bottom-up pipeline: first performing an image segmentation and subsequently classifying the resulting regions into text or non-text ones. Yao \etal~\cite{yao2012detecting} extract regions in the Stroke Width Transform (SWT) domain, proposed earlier for text detection by Epshtein \etal~\cite{Epshtein2010}. Yin \etal~\cite{Yin2013} obtain state-of-the-art performance with a method that prunes the tree of Maximally Stable Extremal Regions (MSER) using the strategy of minimizing regularized variations. The effectiveness of MSER for character candidates detection is also exploited by Chen \etal~\cite{Chen2011} and Novikova \etal~\cite{Novikova2012}, while Neumann \etal~\cite{Neumann2012} propose a region representation derived from MSER where character/non-character classification is done for each possible Extremal Region (ER).

Most of the region-based methods are complemented with a post-processing step where regions assessed to be characters are grouped together into words or text lines. The hierarchical structure of text has been traditionally exploited in a post-processing stage with heuristic rules~\cite{Epshtein2010,Chen2011} usually constrained to search for horizontally aligned text in order to avoid a combinatorial explosion of enumerating all possible text lines. Neumann and Matas~\cite{Neumann2012} introduce an efficient exhaustive search algorithm using heuristic verification functions at different grouping levels (i.e. region pairs, triplets, etc.), but still constrained to horizontal text. Yao \etal~\cite{yao2012detecting} make use of a greedy agglomerative clustering where regions are grouped if their average alignment is under a certain threshold. Yin \etal~\cite{Yin2013} use a self-training distance metric learning algorithm that can learn distance weights and clustering thresholds simultaneously and automatically for text groups detection in a similarity feature space.

\begin{figure*}[t]
\centering
\includegraphics[width=\linewidth]{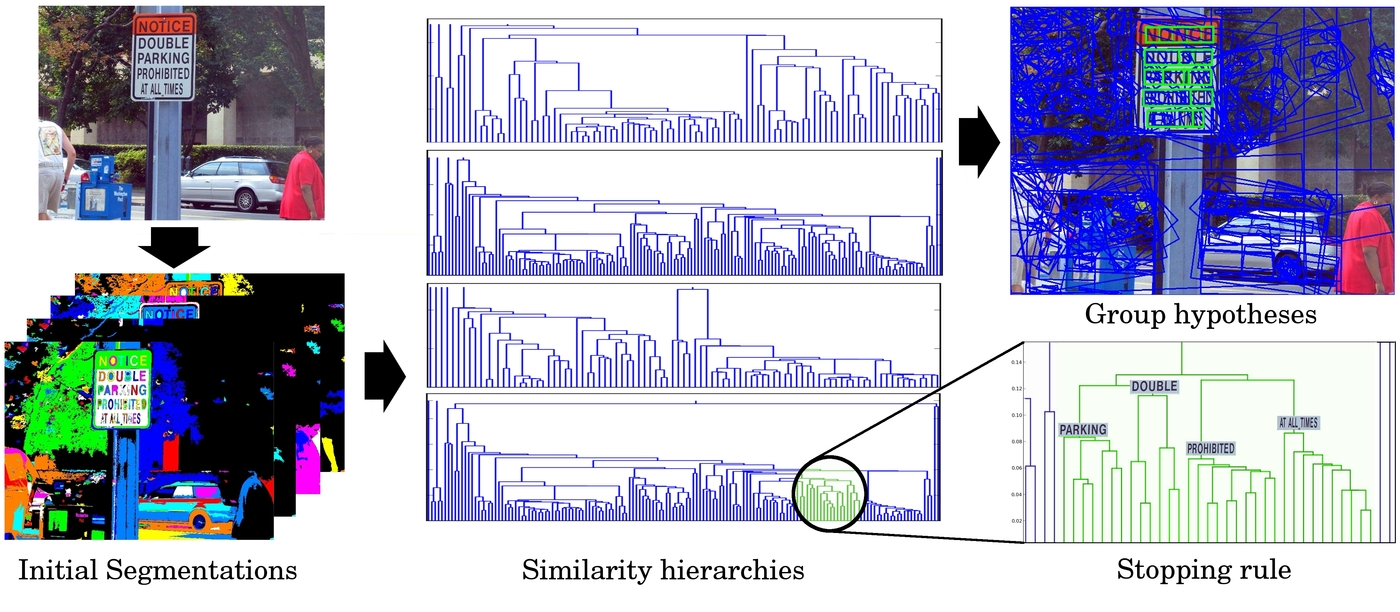}
\caption{A bottom-up agglomerative clustering of individual regions produces a dendrogram in which each node represents a text group hypothesis. Our work focuses on learning the optimal features allowing the generation of pure text groups (comprising only text regions) with high recall, and designing a stopping rule that allows the efficient detection of those groups in a single grouping step.}
\label{fig:hierarchy_dendrogram00}
\end{figure*}

In this paper we present a novel hierarchical approach in which region hierarchies are built efficiently using Single Linkage Clustering in a weighted similarity feature space. The hierarchies are built in different color channels in order to diversify the number of hypotheses and thus increase the maximum theoretical recall. Our method is less heuristic in nature and faster than the greedy algorithm of Yao \etal~\cite{yao2012detecting}, because the number of atomic objects in our clustering analysis is not increased by taking into account all possible region pairs; besides our method uses similarity and not collinearity for grouping. Yin \etal~\cite{Yin2013}, and also our previous work~\cite{Gomez2013}, make use of a two step architecture first doing an automatic clustering analysis in a similarity feature space and then classifying the groups obtained in the first step. The method presented here differs from such approaches in that our agglomerative clustering algorithm integrates a group classifier, acting as a stopping rule, that evaluates the conditional probability for each group in the hierarchy to correspond to a text group in an efficient manner through the use of incrementally computable descriptors. In this sense our work is related with Matas and Zimmerman~\cite{matas2005} region detection algorithm, while the incremental descriptors proposed here are designed to find relevant groups of regions in a similarity dendrogram instead of the detection of individual regions in the component tree of the image. There is also a relationship between our method with the work of Van de Sande \etal~\cite{van2011} and Uijlings \etal~\cite{uijlings2013selective} on using segmentation and grouping as selective search for object recognition. However, our approach is distinct in that their region grouping algorithm agglomerates regions in a class-independent way while our hierarchical clusterings are designed in order to maximize the chances of finding specifically text groups. Thus, our algorithm can be seen as a task-specific selective search.

\section{Hierarchy guided text extraction}
Our hierarchical approach to text extraction involves an initial region decomposition step where non-overlapping atomic text parts are identified. Extracted regions are then grouped together in a bottom-up manner with an agglomerative process guided by their similarity. The agglomerative clustering process produces a dendrogram where each node represents a text group hypothesis. We can then find the branches corresponding to text groups by simply traversing the dendrogram with an efficient stopping rule. Figure~\ref{fig:hierarchy_dendrogram00} shows an example of the main steps of the pipeline.

We make use of the Maximally Stable Extremal Regions (MSER)~\cite{Matas2004} algorithm to get the initial set of low-level regions. MSERs have been extensively used in recent state of the art methods for detecting text character candidates~\cite{Neumann2011,Chen2011,Novikova2012,Gomez2013,Yin2013}. 
Recall in character detection is increased by extracting regions from different single channel projections of the image (i.e. gray, red, green and blue channel). This technique, denoted MSER++~\cite{Neumann2011}, is a way of diversifying the segmentation step in order to maximize the chances of detecting all text regions. 

In the following we address the problem of designing a grouping algorithm exploiting the hierarchical structure of text, in order to detect text regions in a holistic manner. Our solution involves the learning of the optimal clustering feature space for text regions grouping and the design of novel discriminative and probabilistic stopping rules, that allows the efficient detection of text groups in a single clustering step.

\subsection{Optimal clustering feature space}
\label{sec:weighted}

\begin{figure}
  \centering
  \subfloat[]{\includegraphics[width=0.3\linewidth]{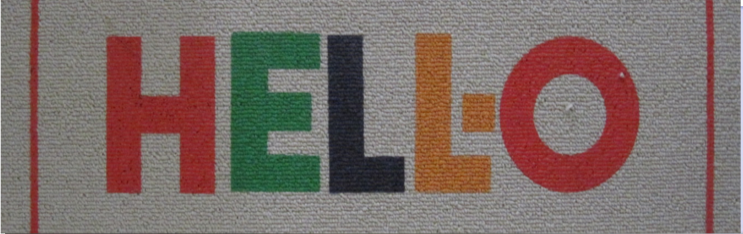}\label{fig:dissimilarity-color}}
  \hspace{0.1cm}
  \subfloat[]{\includegraphics[width=0.3\linewidth]{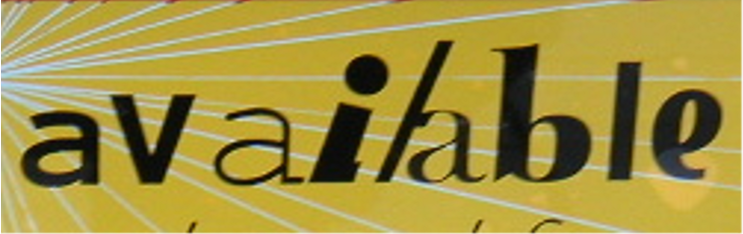}\label{fig:dissimilarity-stroke}}
  \hspace{0.1cm}
  \subfloat[]{\includegraphics[width=0.3\linewidth]{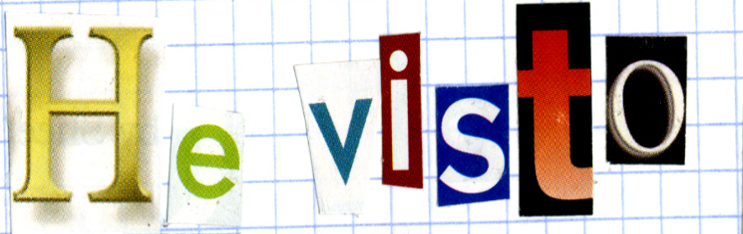}\label{fig:dissimilarity-size}}
  \caption{There is no single best feature for character clustering: Characters in the same word may appear with different color (a), stroke width (b) or sizes (c).}
  \label{fig:dissimilarities}
\end{figure}

It is usually expected that text parts belonging to the same word or text line share similar colors, stroke widths, and sizes.
Although the previous assumption does not always hold (see Figure~\ref{fig:dissimilarities}), in this work we consider that it is possible to weight those simple similarity features obtaining an optimal feature space projection that maximizes the probabilities of finding pure text groups (groups comprising only regions that correspond to text parts) in a Single Linkage Clustering (SLC) dendrogram.

\begin{figure*}[t]
\centering
  \subfloat[]{\includegraphics[width=0.23\linewidth]{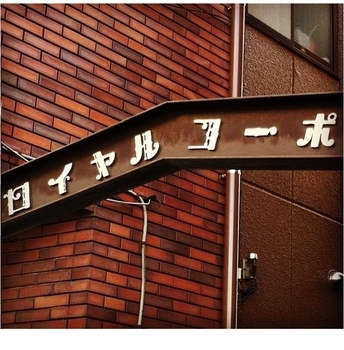}\label{fig:max_recall_a}}
  \hspace{0.1cm}
  \subfloat[]{\includegraphics[width=0.23\linewidth]{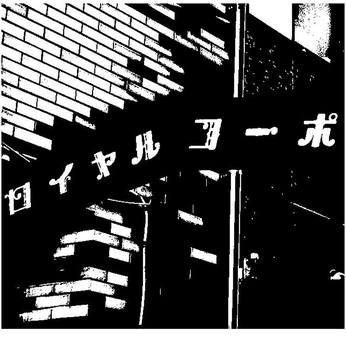}\label{fig:max_recall_b}}
  \hspace{0.1cm}
  \subfloat[]{\includegraphics[width=0.23\linewidth]{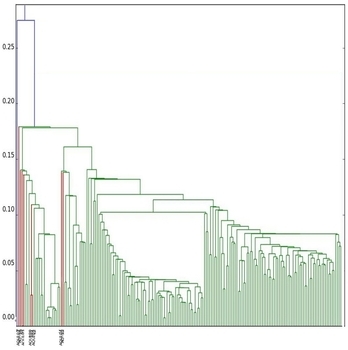}\label{fig:max_recall_c}}
  \hspace{0.1cm}
  \subfloat[]{\includegraphics[width=0.23\linewidth]{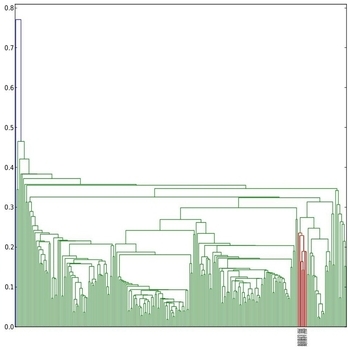}\label{fig:max_recall_d}}
\caption{(a) Scene image, (b) its MSER decomposition, and (c,d) two possible hierarchies built from two different weight configurations, red nodes indicate pure text groupings. The first configuration (c) yields a 28\% text group recall ($\mathcal{TGR}$) while the second (d) achieves 100\% for this particular image. }
\label{fig:max_recall_dendrograms}
\end{figure*}

Let $\mathcal{R}_c$ be the initial set of individual regions extracted with the MSER algorithm from channel $c$. 
We start an agglomerative clustering process, where initially each region $r\in\mathcal{R}_c$ starts in its own cluster and then the closest pair of clusters ($A,B$) is merged iteratively, using the single linkage criterion ($ \min \, \{\, \mathrm{d}(r_a,r_b) : r_a \in A,\, r_b \in B \,\} $), until all regions are clustered together ($C \equiv \mathcal{R}_c$). The distance between two regions $\mathrm{d}(r_a,r_b)$ is calculated as the squared Euclidean distance between their weighted feature vectors, adding a spatial constraint term (the squared Euclidean distance between their centers' coordinates) in order to induce neighbouring regions to merge first: 

\begin{equation} 
  \mathrm{d}(\mathbf{a},\mathbf{b}) = {\sum\limits^D_{i=1}}{(w_i\cdot (a_i - b_i))^2} + \{(x_a-x_b)^2+(y_a-y_b)^2\}
\label{eq:distance}
\end{equation}

\noindent
where we consider the 5-dimensional feature space ($D=5$) comprising the following features: mean gray value of the region, mean gray value in the immediate outer boundary of the region, region's major axis, mean stroke width, and mean of the gradient magnitude at the region's border.

It is worth noting that using the squared Euclidean distance for the spatial term in equation~\ref{eq:distance}, our clustering analysis remains rotation invariant, thus the obtained hierarchy generates the same text group hypotheses independently of the image orientation. For example, rotating the image in Figure~\ref{fig:max_recall_a} by any degree would produce exactly the same dendrograms shown in Figures~\ref{fig:max_recall_c} and \ref{fig:max_recall_d}.This is intentional as we want our method to be capable of detecting text in arbitrary orientations. In this way, our algorithm deals naturally with arbitrary oriented text without using any heuristic assumption or threshold.

Given a possible set of weights $\mathbf{w}$, SLC produces a dendrogram $D_w$ where each node $H \in {D_w}$ is a subset of $\mathcal{R}_c$ and represents a text group hypothesis. The text group recall ($\mathcal{TGR}$) represents the ability of a particular weighting configuration to produce pure text groupings (comprising only text regions) corresponding to words or text lines in the ground-truth. Figure~\ref{fig:max_recall_dendrograms} shows an example of how different weight configurations lead to different text group recall.

Given a set of hypotheses $H \in D_w$, and a set of ground-truth text-group objects (i.e. words and text-lines) $G \in GT$, $\mathcal{TGR}$ is defined as:

\begin{equation}
 \begin{split}
   \mathcal{TGR}(D_w,GT) &= {1\over{|GT|}} \sum_{G \in GT} (\max_{H \in D_w}{{|H|}\over{|G|}} \\
   &  \mid \forall \, {r_h}\in{H} \, \exists \, {r_g}\in{G} \mid m_r(r_h,r_g) > 0.9 )
 \end{split}
\end{equation}

\noindent
where $|\cdot|$ indicates cardinality of a set, and $m_r(\cdot,\cdot)$ is the overlap ratio between two regions. Thus, for each text-group in the ground-truth ($G \in GT$) we look for the largest group hypothesis in the dedrogram ($H \in D_w$) such that all regions in $H$ ($r_h \in H$) match with regions in the ground-truth group (${r_g}\in{G}$). 

Our optimization problem for learning the optimal clustering feature space is defined as finding the set of weights $w_{opt}$ that maximise $\mathcal{TGR}$:

\begin{equation}
  w_{opt} = \argmax_w \{ \mathcal{TGR}(D_w,GT) \}
\label{eq:argmax}
\end{equation}

As discussed before, there is no single best way to define similarity between text parts, hence there is no single best set of weights for our strategy, instead missing groups under a particular configuration may be potentially found under another.
An alternative to using a single feature space would be to diversify our clustering strategy, adding more hypotheses to the system by building different hierarchies obtained from different weight configurations (similarly to what we do with different color channels). 

\begin{figure}
\centering
\includegraphics[width=\linewidth]{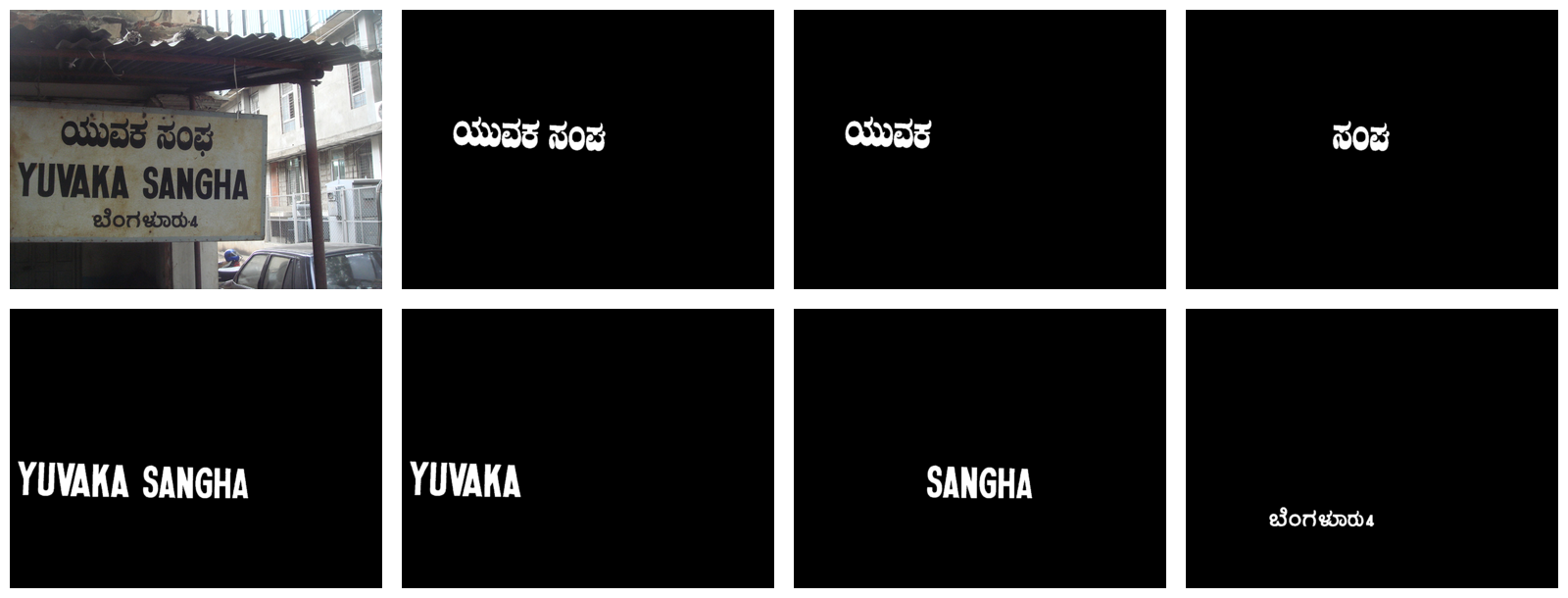}
\caption{Our training set is assembled by manually separating the pixel level ground-truth of train images into all possible text groups (lines and words).}
\label{fig:train_sample}
\end{figure}

At training time, we use grid search strategy over the weights parameter space in order to solve equation~\ref{eq:argmax} for our training dataset. We assembled a mixed set of training examples using the MSRRC and ICDAR training sets. The MSRRC training set contains 167 images and the ICDAR training set 229. We have manually separated all text-lines and words in the ground truth data of these images, giving rise to 1611 examples of text groups. Figure~\ref{fig:train_sample} shows the group examples extracted from one of the training set images.

The optimized weights $w_{opt}$, obtained with grid search by maximizing the text group recall using equation~\ref{eq:argmax}, yield a cross-validated $\mathcal{TGR}$ of 0.87 in the training set, and are:
  {$w_1 = 0.65$ (for intensity mean)}
  , {$w_2 = 0.65$ (for outer boundary intensity mean)}
  , {$w_3 = 0.49$ (for border gradient mean)}
  , {$w_4 = 0.67$ (for diameter)}
  , {$w_5 = 0.91$ (for stroke width mean)}.

  As diversification strategy, after an optimized set of weights is obtained we subsequently remove from the training set the groups that have been detected, and then search again for new optimal weights $(w_{opt_{2}}, \dots, w_{opt_{n}})$ with the remaining groups. We evaluate this diversification strategy in section~\ref{sec:experiments}.

At test time, each of the optimal weight configurations is used to generate a dendrogram where each node represents a text group hypothesis. Selecting the branches corresponding to text groups is done by traversing the dendrogram with an efficient stopping rule.

\subsection{Discriminative and Probabilistic Stopping Rules}

Given a dendrogram representing a set of text groups hypotheses from the SLC algorithm, we need a strategy to determine the partition of the data that best fits our objective of finding pure text groups. A rule to decide the best partition in a Hierarchical Clustering is known as a \emph{stopping rule} because it can be seen as stopping the agglomerative process.
Differently from standard clustering stopping rules here we do not expect to obtain a full partition of regions in $\mathcal{R}_c$. In fact we do not even know if there are any text clusters at all in $\mathcal{R}_c$. Moreover, in our case we have a quite clear model for the kind of groups sought, corresponding to text. These particularities motivate the next contribution of this paper. We propose a stopping rule, to select a subset of meaningful clusters in a given dendrogram $D_w$, comprising the combination of the following two elements:

\begin{itemize}
  \item{A text group discriminative classifier.}
  \item{A probabilistic measure for hierarchical clustering validity assessment~\cite{Cao2004}.}
\end{itemize}

\subsubsection{Discriminative Text Group Classifier}
\label{sec:discriminative}

The first part of our stopping rule takes advantage of supervised learning, building a discriminative classification model $\mathcal{F}$ in a group-level feature space. Thus, given a group hypothesis $H$ and its feature vector $\mathbf{h}$, our stopping rule will accept $H$ only if $\mathcal{F}(\mathbf{h}) = 1$. We use a Real AdaBoost classifier~\cite{schapire1999improved} with decision stumps.
Our group-level features originate from three different types: 1) Group intra-similarity statistics, since we expect to see regions in the same word having low variation in color, stroke width, and size; 2) Shape similarity of participating regions, in order to discriminate repetitive patterns, such as bricks or windows in a building, which tend to be confused with text; 3) Collinearity and equidistance features, measure the text-like structure of text groups by using statistics of the 2-D Minimum Spanning Tree (MST) built with their regions centers. The list of used features is as follows:

\begin{itemize}
      \item{FG intensities standard deviation.}
      \item{BG intensities standard deviation.}
      \item{Major axis coefficient of variation.}
      \item{Stroke widths~\cite{Chen2011} coefficient of variation.}
      \item{Mean gradient standard deviation.}
      \item{Aspect ratios coefficient of variation.}
      \item{Hu's invariant moments~\cite{hu1962} average Euclidean distance.}
      \item{Convex hull compactness~\cite{Neumann2012} mean and standard deviation.}
      \item{Convexity defects coefficient of variation.}
      \item{MST angles mean and standard deviation.}
      \item{MST edge widths coefficient of variation.}
      \item{MST edge distances mean vs. diameters mean ratio.}
\end{itemize}

The AdaBoost classifier is trained using the same training set described in section~\ref{sec:weighted}. We have two sources of positive samples: 1) Using each GT group as if it were the output of the region decomposition step; 2) we run MSER and SLC ($w_{opt}$) against a train image and use as positive samples those pure-text groups in the SLC tree with more than 80\% match with a GT group. From the same tree we extract negative examples as nodes with 0 matchings. This gives us around 3k positive and 15k negative samples. We balance the positive and negative data and train a first classifier that is used to select 100 hard negatives that are used to re-train and improve accuracy.

\subsubsection{Incrementally computable descriptors}

Since at test time we have to calculate the group level features at each node of the similarity hierarchy, it is important that they are fast to compute. We take advantage of the inclusion relation of the dendrogram's nodes in order to make such features incrementally computable when possible. This allows us to compute the probability of each possible group of regions to be a text group without affecting the overall time complexity of our algorithm.

Group level features consisting of simple statistics over individual region features (e.g. diameters, strokes, intensity, etc.) can be incrementally computed straightforwardly with a few arithmetic operations and so have a constant complexity $\mathcal{O}(1)$.

Regarding the MST based features, an incremental algorithm (i.e. propagating the MST of children nodes to their parent) computing the MST on each node of the dendrogram takes $\mathcal{O}(n \times {\log}^2{n})$ in the worst case. Although this complexity is much lower than the $\mathcal{O}(n^2)$ complexity of the SLC step and thus does not affect the overall complexity of the algorithm, this has noticeable impact in run time. For this reason we add an heuristic rule on the maximum size of valid clusters: clusters with more than a certain number of regions are immediately discarded and there is no need to compute their features. By taking the length of the largest text line in the MSRRC training set (50) as the maximum cluster size, the run-time growth due to the features calculation in our algorithm is negligible and the obtained results are not affected at all. 

\subsubsection{Probabilistic cluster meaningfulness estimation}

\begin{figure}
\centering
\includegraphics[width=\linewidth]{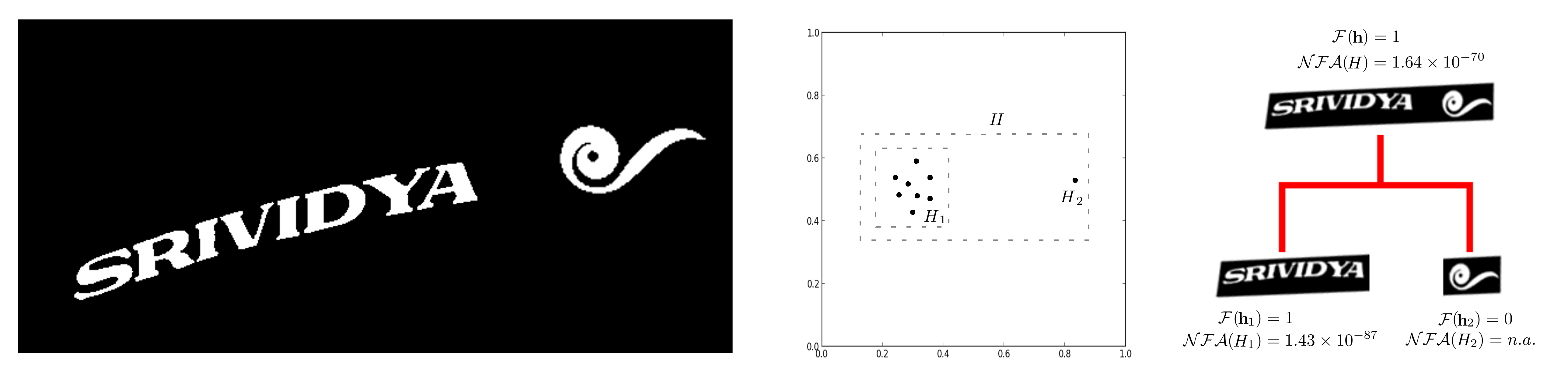}
\caption{A node in a similarity dendrogram consisting in a correctly detected word ($H_1$) is merged with a cluster consisting of a single region outlier ($H_2$). Our stopping rule will not consider valid the resulting cluster $H = \{H_1 \cup H_2\}$ although the classifier has labelled it as a text group ($\mathcal{F}(\mathbf{h}) = 1$) because $\mathcal{NFA}(H)$ is larger than $\mathcal{NFA}(H_1)$. The scatter plot simulates the arrangement of the feature vectors of the regions forming $H_1$, $H_2$, and $H$ in the similarity feature space.}
\label{fig:nfa_rule}
\end{figure}

The way our classifier $\mathcal{F}$ is designed may eventually make the discriminative stopping rule to accept groups with outliers. For example, Figure~\ref{fig:nfa_rule} shows the situation where a node of the dendrogram consisting in a correctly detected word is merged with a (character like) region which is not part of the text group (outlier). In order to increase the discriminative power of our \emph{stopping rule} in such situations, we make use of a probabilistic measure of cluster meaningfulness~\cite{Desolneux2003,Cao2004}. This probabilistic measure, also used for text detection in our previous work~\cite{Gomez2013}, provides us with a way to compare clusters' qualities in order to decide if a given node in the dendrogram is a better text candidate than its children. 

The Number of False Alarms ($\mathcal{NFA}$)~\cite{Desolneux2003,Cao2004}, based on the principle on non-accidentalness, measures the meaningfulness of a particular group of regions in $\mathcal{R}_c$ by quantifying
how the distribution of their features deviates from randomness.
Consider that there are $n$ regions in $\mathcal{R}_c$ and that a particular group hypothesis $H$ of $k$ of them have a feature in common. Assuming that the observed quality has been distributed randomly and uniformly across all regions in $\mathcal{R}_c$, the probability that the observed distribution for $H$ is a random realisation of this uniform process is given by the tail of the binomial distribution:

\begin{equation}
  \mathcal{NFA}(G) = \mathcal{B}_G(k,n,p) = \sum_{i=k}^{n} \binom{n}{i} p^i (1-p)^{n-i}
\label{eq:nfa}
\end{equation}

\noindent
where $p$ is the probability of a single object having the aforementioned feature. The lower the $\mathcal{NFA}$ is, the more meaningful the group is.

We make use of this metric in each node of a dendrogram $D_w$ to assess the meaningfulness of all produced grouping hypotheses. We calculate (\ref{eq:nfa}) for each possible group hypothesis $H$ using as $p$ the ratio of the volume defined by the distribution of the feature vectors of the comprising regions ($h \in H$) with respect to the total volume of the $5-D$ feature space defined in Section~\ref{sec:weighted}.

Our stopping rule is defined recursively in order to accept a particular hypothesis $H$ as a valid group \emph{iif} its classifier predicted label is "text" ($\mathcal{F}(\mathbf{h}) = 1$) and its meaningfulness measure is higher than the respective meaningfulness measures of every successor $A$ and every ancestor $B$ labelled as text, i.e. the following inequalities hold:

\begin{equation}
  \mathcal{NFA}(H) < \mathcal{NFA}(A), \forall A \in successors(H) \mid \mathcal{F}(A)=1
\end{equation}
\begin{equation}
  \mathcal{NFA}(H) < \mathcal{NFA}(B), \forall B \in ancestors(H) \mid \mathcal{F}(B)=1
\end{equation}

Notice that by using this criteria no region is allowed to belong to more than one text group at the same time. The clustering analysis is done without specifying any parameter or cut-off value and without making any assumption on the number of meaningful clusters, but merely comparing the values of (\ref{eq:nfa}) at each node in the dendrogram for which the discriminative classifier label is "text" ($\mathcal{F}(H)=1$).
See Figure~\ref{fig:nfa_rule} for an example on how this stopping rule is able to detect outliers. As a side effect, the stopping rule is also able to correctly separate different words in a text line.

\setlength{\tabcolsep}{4pt}
\begin{table*}[t]
  \centering
  \caption{Segmentation results (a) and Precision-Recall curves (b) comparing different variants of our method.}
  \subfloat[]{\baselinetab}
  \hspace{0.6cm}
  \subfloat[]{\includegraphics[width=0.34\linewidth]{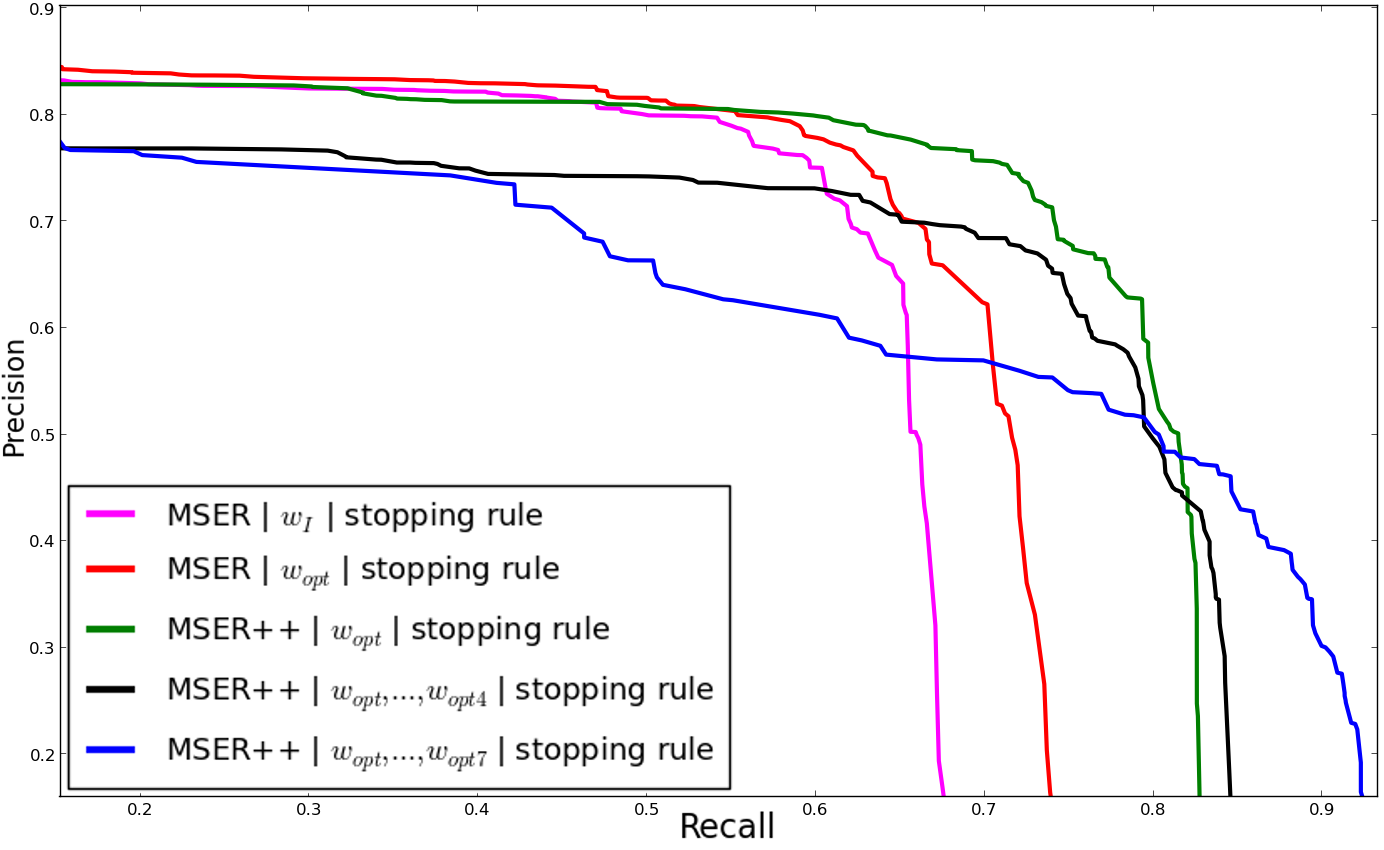}\label{fig:pr}}
  \label{table:results_msrrc_diversification}
\end{table*}
\setlength{\tabcolsep}{1.4pt}

\begin{figure*}
\centering
\includegraphics[width=\linewidth]{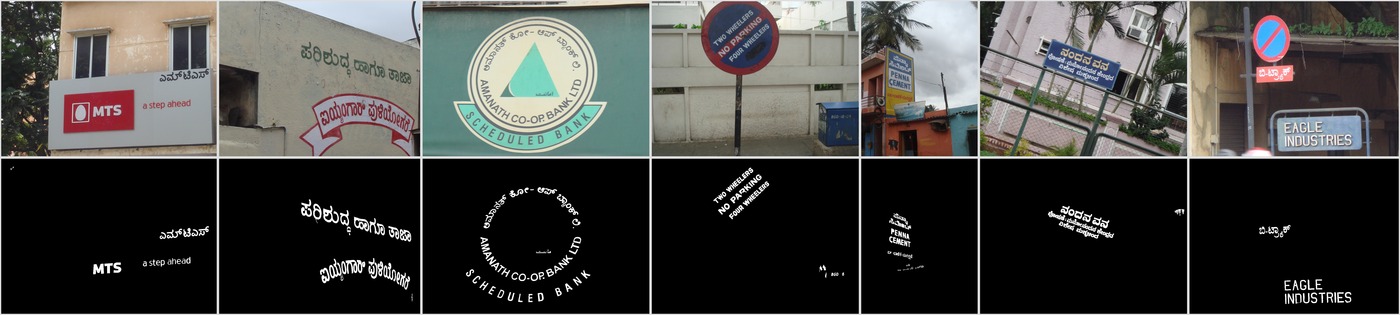}
\caption{Qualitative segmentation results on the MSRRC 2013 dataset.}
\label{fig:msrrc_ok}
\end{figure*}

\begin{figure*}
\centering
\includegraphics[width=\linewidth]{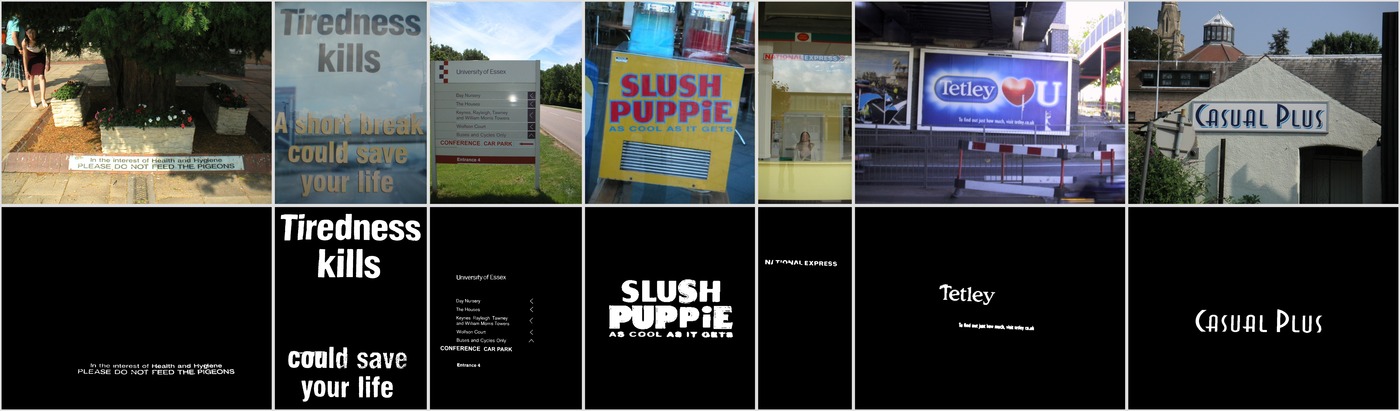}
\caption{Qualitative segmentation results on the ICDAR dataset.}
\label{fig:icdar_ok}
\end{figure*}

At this point, applying the method described so far our algorithm is able to produce results for the scene text segmentation task. The segmentation task is evaluated at pixel-level, this is the algorithm must provide a binary image where white pixels correspond to text and black pixels to background. All segmentation results given in section~\ref{sec:experiments} are obtained with this algorithm, trained with a single mixed dataset and without any further post-processing, by setting to white the pixels corresponding to the detected text groups. Figures~\ref{fig:msrrc_ok}, \ref{fig:kaist_ok}, and \ref{fig:icdar_ok} show segmentation results of our method in different datasets.

\subsection{From Pixel Level Segmentation to Bounding Box Localization}

In order to evaluate our method in the text localization task we extend our method with a simple post-processing operation in order to obtain word and text line bounding boxes depending on the semantic level ground truth information is defined (e.g. words in the case of ICDAR and MSRRC datasets, lines in the case of the MSRA-TD500 dataset). This is because the text groups detected by our stopping rule may correspond indistinctly to words, lines, or even paragraphs in some cases, depending on the particular typography and layout of the detected text.

First of all, region groups selected as text by our stopping rule in the different dendrograms are combined in a procedure that serves to de-duplicate repeated groups (e.g. the same group may potentially be found in several channels or weights configurations) and to merge collinear groups that may have been detected by chunks. Two given text groups are merged if they are collinear, and their relative distance and height ratio are under thresholds learned during training.

After that, if needed by the granularity of the ground-truth level, we split resulted text lines into words by considering as word boundaries all spaces between regions with a larger distance than a certain threshold, learned during training, proportional to the group's average inter-region distance.

\section{Experiments}
\label{sec:experiments}

The proposed method has been evaluated on three multi-script and arbitrary oriented text datasets and one English-only dataset for the tasks of text extraction and localization. All the segmentation evaluation is done at the pixel level, i.e. precision $p$ and recall $r$ are defined as $p = |E \cap T| / |E|$ and $r = |E \cap T| / |T|$, where $E$ is the set of pixels labelled as text and $T$ is the set of pixels corresponding to text in the ground truth. The localization results are evaluated with different frameworks depending on the dataset. The ICDAR~\cite{karatzas2013icdar,Lucas2003} and MSCCR~\cite{kumar2013multi} datasets have ground-truth defined at the word level and the proposed evaluation framework is the one of Wolf and Jolion~\cite{wolf2006}. The MSRA-TD500~\cite{yao2012detecting} has ground-truth defined at the line level and uses it's own evaluation framework. We use the standard evaluation frameworks for each dataset to be able to compare with the state of the art.


\subsection{Baseline analysis}

We have evaluated different variants of our method in order to assess the contribution of each of the proposed techniques. This baseline analysis is performed in the MSRRC test set. The baseline method is configured by setting all weights to $1$ ($w_I$) and accepting all group hypotheses which are labelled as text by the classifier ($\mathcal{F}(H)=1$). We compare this baseline with the variants making use of the learned optimal weights $w_{opt}$, and with including the meaningfulness criteria to our \emph{stopping rule}. Finally we have evaluated the impact of different diversification strategies to the initial segmentation, both in the number of image channels (MSER vs. MSER++), and in the number of weight configurations by adding a variable number of optimal weighted configurations $w_{opt_2}, \dots, w_{opt_n}$ into the system. 
Table~\ref{tab:baselinetab} shows segmentation results of our method in the MSRRC 2013 test set comparing different variants of our method and different diversification strategies. The table includes also two variants of our previously published work~\cite{Gomez2013} for comparison. We chose to use the MSRRC dataset for this analysis as it is representative of the targeted scenario of multi-script and arbitrarily oriented text. Figure~\ref{fig:pr} plots the Precision-Recall curves, obtained by varying the acceptance threshold of the discriminative classifier in the stopping rule, for the five top scoring variations in Table~\ref{tab:baselinetab}.

From the obtained results we can see that the optimized weights $w_{opt}$ have a noticeable impact in the method recall, while the \emph{stopping rule} leads to a considerable increase in precision without any recall deterioration.
Regarding diversification, if one wants to maximize the harmonic mean between precision and recall, the use of MSER++ is well justified even though it produces a slight precision drop. However, examining the effect of further diversification using more optimal weighting configurations, we can see that the obtained gain in recall by adding more hypotheses does not help improving the f-score as it produces a significant precision deterioration. Such a diversification strategy should be considered only if one wants to maximize the system's recall.

\subsection{Multi-script and arbitrary oriented scene text extraction}
\label{sec:experimets_msrrc}

\begin{figure*}
\centering
\includegraphics[width=\linewidth]{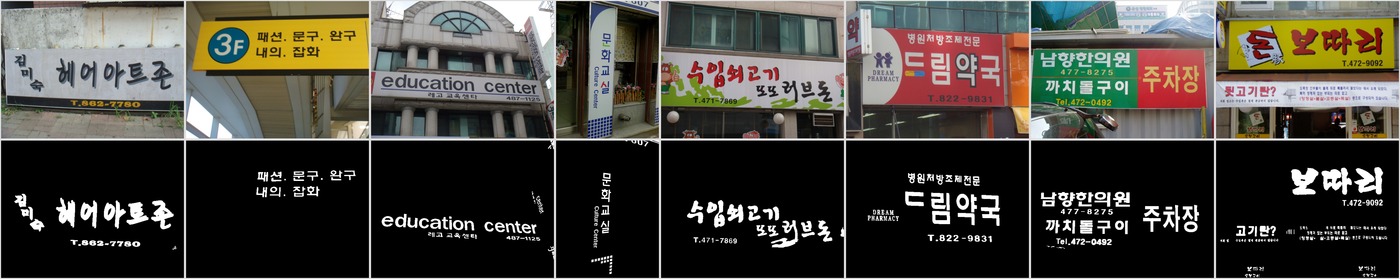}
\caption{Qualitative segmentation results on the KAIST dataset.}
\label{fig:kaist_ok}
\end{figure*}

\begin{figure*}
\centering
\includegraphics[width=\linewidth]{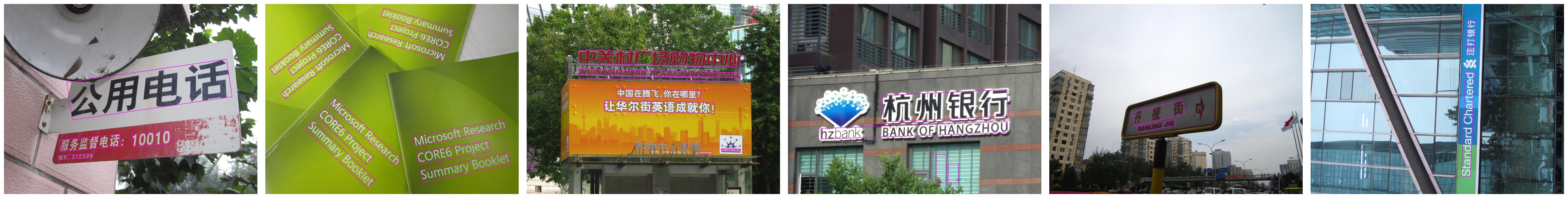}
\caption{Qualitative localization results on the MSRA-TD500 dataset.}
\label{fig:msra_localization}
\end{figure*}

We evaluate our method in three standard multi-script and arbitrary oriented text datasets. The MSRA-TD500 dataset~\cite{Yao2012} does not have pixel-level segmentation ground truth and thus we only evaluate on it for the text localization task, while in the MSRRC~\cite{kumar2013multi} and KAIST~\cite{Lee2010} datasets evaluation is done for both segmentation and localization tasks.

The MSRA-TD500 dataset~\cite{Yao2012} contains arbitrary oriented text in both English and Chinese languages. The dataset contains 500 images in total, with varying resolutions from $1296\times864$ to $1920\times1280$. The evaluation for text localization is done as proposed in~\cite{Yao2012} using minimum area rectangles. For an estimated minimum area rectangle $D$ to be considered a true positive, it is required to find a ground truth rectangle $G$ such that:

\begin{equation*}
  {A(D' \cap G')}/{A(D' \cup G')} > 0.5  ,  abs(\alpha_D - \alpha_G) < {\pi}/{8}
\end{equation*}

where $D'$ and $G'$ are the axis oriented versions of $D$ and $G$, $A(D' \cap G')$ and $A(D' \cup G')$ are respectively the area of their intersection and union, and $\alpha_D$ and $\alpha_G$ their rotation angles. The definitions of precision $p$ and recall $r$ are: $p = |TP|/|E|$, $r = |TP|/|T |$ where $TP$ is the set of true positive detections while $E$ and $T$ are the sets of estimated rectangles and ground truth rectangles. Table~\ref{table:results_msra} compares our results with other state of the art methods on the MSRA-TD500 dataset and Figure~\ref{fig:msra_localization} show qualitative localization results.

\setlength{\tabcolsep}{4pt}
\begin{table}[h]
\begin{center}
\caption{Localization results in the MSRA-TD500 dataset.}
\label{table:results_msra}
\begin{tabular}{l c c c}
\hline\noalign{\smallskip}
Method & Precision & Recall & F-score\\
\noalign{\smallskip}
\hline
\noalign{\smallskip}
\cellcolor{blue!25}\textbf{Ours} & \cellcolor{blue!25}\textbf{0.69} & \cellcolor{blue!25}0.54 & \cellcolor{blue!25}\textbf{0.61}\\
       TD-Mixture~\cite{Yao2012} & 0.63 & \textbf{0.63} & 0.60\\
Gomez \& Karatzas \cite{Gomez2013}& 0.58 & 0.54 & 0.56\\
TD-ICDAR~\cite{Yao2012}& 0.53 & 0.52 & 0.54 \\
Epshtein \etal~\cite{Epshtein2010}& 0.25 & 0.25 & 0.25\\
Chen \etal~\cite{Chen2004}& 0.05 & 0.05 & 0.05\\
\hline
\end{tabular}
\end{center}
\end{table}
\setlength{\tabcolsep}{1.4pt}

The MSRRC dataset~\cite{kumar2013multi} comprises 334 camera-captured scene images, 167 in the training and 167 in the test set respectively, with sizes around $1.2MP$ for text localization and segmentation tasks. It covers Latin, Chinese, Kannada, and Devanagari scripts, and includes text with multiple orientations. Tables~\ref{table:results_msrrc} and ~\ref{table:results_msrrc2} compares our results with the participants in the segmentation and localization tasks of the 2013 Multi-script Robust Reading Competition, while Figure~\ref{fig:msrrc_ok} show examples of qualitative results. The average run-time of our algorithm in this dataset is 1.67 seconds per image on a standard PC. 

\begin{figure}
\centering
\includegraphics[width=0.8\linewidth]{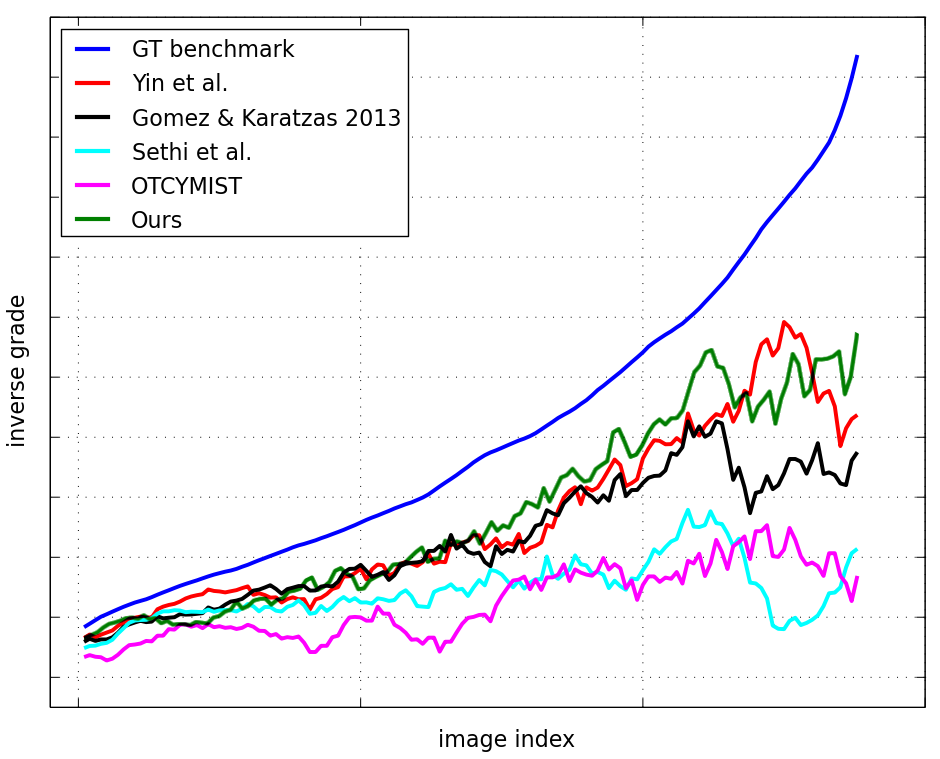}
\caption{Inverse grade curves of different methods in the MSRRC dataset~\cite{kumar2013multi}.}
\label{fig:inverse}
\end{figure}

\setlength{\tabcolsep}{4pt}
\begin{table}[h]
\begin{center}
\caption{Segmentation results in the MSRRC 2013 dataset.}
\label{table:results_msrrc}
\begin{tabular}{l c c c}
\hline\noalign{\smallskip}
Method & Precision & Recall & F-score\\
\noalign{\smallskip}
\hline
\noalign{\smallskip}
\cellcolor{blue!25}\textbf{Ours} & \cellcolor{blue!25}\textbf{0.75} & \cellcolor{blue!25}\textbf{0.71} & \cellcolor{blue!25}\textbf{0.73}\\
Yin \etal  \cite{Yin2013,kumar2013multi} & 0.71 & 0.67 & 0.69\\
Gomez \& Karatzas \cite{Gomez2013,kumar2013multi}& 0.64 & 0.58 & 0.61\\
Sethi \etal \cite{kumar2013multi}& 0.33 & 0.72 & 0.45 \\
OTCYMIST \cite{kumar2012otcymist,kumar2013multi}& 0.50 & 0.29 & 0.37\\
\hline
\end{tabular}
\end{center}
\end{table}
\setlength{\tabcolsep}{1.4pt}

\setlength{\tabcolsep}{4pt}
\begin{table}[h]
\begin{center}
\caption{Localization results in the MSRRC 2013 dataset.}
\label{table:results_msrrc2}
\begin{tabular}{l c c c}
\hline\noalign{\smallskip}
Method & Precision & Recall & F-score\\
\noalign{\smallskip}
\hline
\noalign{\smallskip}
\cellcolor{blue!25}\textbf{Ours} & \cellcolor{blue!25}0.63 & \cellcolor{blue!25}\textbf{0.54} & \cellcolor{blue!25}\textbf{0.58}\\
Yin \etal  \cite{Yin2013,kumar2013multi} & \textbf{0.64} & 0.42 & 0.51\\
\hline
\end{tabular}
\end{center}
\end{table}
\setlength{\tabcolsep}{1.4pt}

Figure~\ref{fig:inverse} show the inverse grade curves of different methods in the MSRRC dataset. The inverse grade curve plots the f-score divided by the ratio of text pixels for each image, and inversely sorts these values by the amount of text pixels, thus larger values in the x-axis correspond to images with less text. As can be seen our curve is the nearest to follow the ground-truth benchmark curve.

\begin{figure*}
\centering
\includegraphics[width=\linewidth]{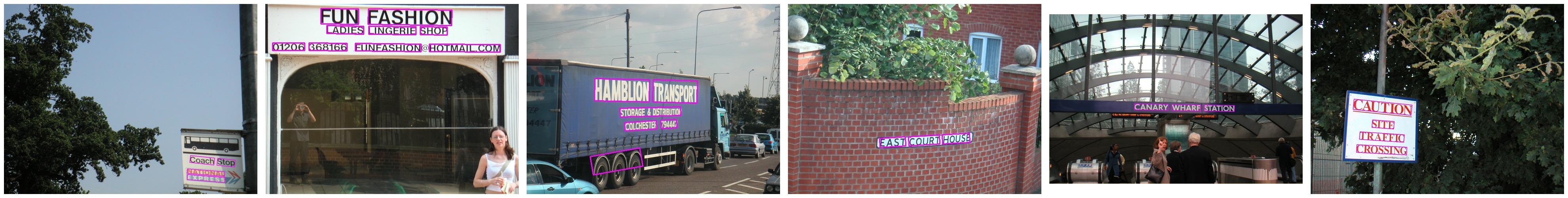}
\caption{Qualitative localization results on the ICDAR 2013 dataset.}
\label{fig:icdar_localization}
\end{figure*}

The KAIST dataset~\cite{Lee2010} comprises 3000 natural scene images, with a resolution of $640\times480 pixels$, categorized according to the language of the scene text captured: Korean, English, and Mixed (Korean + English). For our experiments we use the subset of 800 images corresponding to the Mixed subset in accordance to other reported results. Quantitative results are given in Table~\ref{table:results_kaist} while a set of example qualitative results are shown in Figure~\ref{fig:kaist_ok}. Our method archives a $0.77$ f-score in the localization task, with a $0.71$ precision and $0.83$ recall, not comparable to other methods as being the first to report localization results for this dataset. The average run-time of our algorithm in this dataset is 0.5 seconds per image on a standard PC.

\setlength{\tabcolsep}{4pt}
\begin{table}[h]
\begin{center}
\caption{Segmentation results in the KAIST dataset.}
\label{table:results_kaist}
\begin{tabular}{l c c c}
\hline\noalign{\smallskip}
Method & Precision & Recall & F-score\\
\noalign{\smallskip}
\hline
\noalign{\smallskip}
\cellcolor{blue!25}\textbf{Ours} & \cellcolor{blue!25}\textbf{0.67} & \cellcolor{blue!25}\textbf{0.89} & \cellcolor{blue!25}\textbf{0.76}\\
Gomez \& Karatzas \cite{Gomez2013}& 0.66 & 0.78 & 0.71 \\
Lee \etal \cite{Lee2010} & 0.69 & 0.60 & 0.64 \\
OTCYMIST \cite{kumar2012otcymist} & 0.52 & 0.61 & 0.56 \\
\hline
\end{tabular}
\end{center}
\end{table}
\setlength{\tabcolsep}{1.4pt}

The interpretation of the high increase in recall observed in the KAIST dataset compared to the obtained in MSRRC follows the fact that in KAIST dataset small text characters are not labelled in the groud-truth. These small text components are in general the ones more difficult to detect. On the other hand, in some cases precision suffers when such small text is correctly detected as it counts as false positive.

\subsection{English horizontal scene text extraction}

The proposed method has been evaluated on the ICDAR2013 Robust Reading Dataset~\cite{karatzas2013icdar}. The ICDAR2013 dataset contains 462 images, of which 229 comprise the training set and 233 images the test set. Table~\ref{table:results_icdar} compares the results of our method with the participants in the 2013 ICDAR Robust Reading Competition for the task of text segmentation. The average run-time of our algorithm in this dataset is 1.78 seconds per image on a standard PC. 

\setlength{\tabcolsep}{4pt}
\begin{table}[h]
\begin{center}
\caption{Segmentation results in the ICDAR Robust Reading Competition 2013 dataset.}
\label{table:results_icdar}
\begin{tabular}{l c c c}
\hline\noalign{\smallskip}
Method & Precision & Recall & F-score\\
\noalign{\smallskip}
\hline
\noalign{\smallskip}
I2R NUS FAR \cite{karatzas2013icdar} * & \textbf{0.82} & \textbf{0.75} & \textbf{0.78}\\
I2R NUS \cite{karatzas2013icdar} * & 0.79 & 0.73 & 0.76\\
 \cellcolor{blue!25}\textbf{Ours} & \cellcolor{blue!25}0.74 & \cellcolor{blue!25}0.71 & \cellcolor{blue!25}0.73\\
USTB FuStar \cite{Yin2013,karatzas2013icdar}& 0.74 & 0.70 & 0.72\\
Text Detection \cite{karatzas2013icdar}& 0.76 & 0.65 & 0.70 \\
NSTextractor \cite{karatzas2013icdar}& 0.76 & 0.61 & 0.68 \\
NSTsegmentator \cite{karatzas2013icdar}& 0.64 & 0.68 & 0.66\\
Gomez \& Karatzas 2013 \cite{Gomez2013}& 0.63 & 0.59 & 0.61\\
OTCYMIST \cite{kumar2012otcymist,karatzas2013icdar}& 0.46 & 0.59 & 0.52\\
\hline
\end{tabular}
\end{center}
\end{table}
\setlength{\tabcolsep}{1.4pt}

\setlength{\tabcolsep}{4pt}
\begin{table}[h]
\begin{center}
\caption{Localization results in the ICDAR Robust Reading Competition 2013 dataset.}
\label{table:results_icdar2}
\begin{tabular}{l c c c}
\hline\noalign{\smallskip}
Method & Precision & Recall & F-score\\
\noalign{\smallskip}
\hline
\noalign{\smallskip}
USTB TexStar \cite{karatzas2013icdar,Yin2013} & 0.88 & 0.66 & 0.76\\
TextSpotter  \cite{karatzas2013icdar,Neumann2012}& 0.88 & 0.65 & 0.74\\
CASIA NLPR \cite{karatzas2013icdar} & 0.79 & 0.68 & 0.73\\
 \cellcolor{blue!25}\textbf{Ours} & \cellcolor{blue!25}0.78 & \cellcolor{blue!25}0.67 & \cellcolor{blue!25}0.72\\
Text detector CASIA \cite{karatzas2013icdar}& 0.85 & 0.63 & 0.72\\
I2R NUS FAR \cite{karatzas2013icdar} * & 0.75 & 0.69 & 0.72\\
I2R NUS  \cite{karatzas2013icdar} * &0.73 & 0.66 & 0.69\\
TH-TextLoc  \cite{karatzas2013icdar}&0.70 & 0.65 & 0.67\\
Text Detection \cite{karatzas2013icdar}& 0.74 & 0.53 & 0.62\\
Baseline \cite{karatzas2013icdar}& 0.61 & 0.35 & 0.44\\
Inkam \cite{karatzas2013icdar}& 0.31 & 0.35 & 0.33\\
\hline
\end{tabular}
\end{center}
\end{table}
\setlength{\tabcolsep}{1.4pt}

\begin{figure*}
\centering
\includegraphics[width=\linewidth]{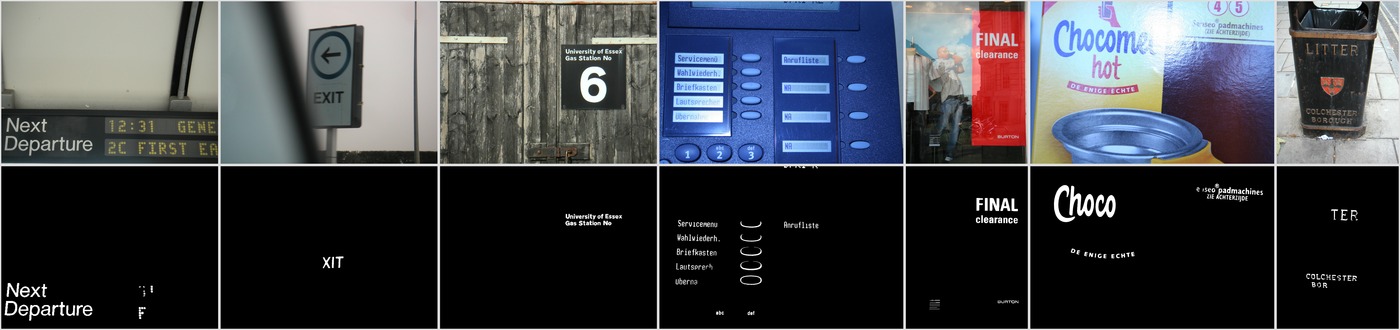}
\caption{Common errors in the ICDAR dataset include false positives due to repetitive patterns and missing text in images with strong highlights, degraded text, or individual characters.}
\label{fig:icdar_fails}
\end{figure*}

As can be seen in Tables~\ref{table:results_icdar} and \ref{table:results_icdar2} our method produces competitive results although it does not perform better than the winner methods in the last ICDAR competition. This has a coherent interpretation as we aim for the highest generality of our method, addressing the unconstrained problem of detecting text irrespective of its language, script, and orientation. Contrary to our method, most methods listed in Tables~\ref{table:results_icdar} and \ref{table:results_icdar2} have been trained explicitly for horizontally aligned English text and address only this particular scenario. For example the TextSpotter \cite{Neumann2012} method is reported to be specifically designed to detect only English and horizontal text. The USTB TextStar~\cite{Yin2013} is a multi-script method but has two different variants for horizontal and arbitrary oriented text, while scoring first in the ICDAR dataset, Tables~\ref{table:results_msrrc} and ~\ref{table:results_msrrc2} show that our method is superior in other scenarios. Methods marked with an asterisk in Tables~\ref{table:results_icdar} and \ref{table:results_icdar2} have not been published.  

Finally, we evaluate our method in the ICDAR2003 dataset~\cite{Lucas2003}. This is a slightly different version of the ICDAR dataset, with almost the same images but proposing a distinct evaluation framework. Table~\ref{table:results_icdar3} compare our results in the ICDAR2003 dataset with other state of the art methods. It is important to notice that some of the top scoring methods in this table have been evaluated in the MSRA-TD500 arbitrary oriented text dataset with a much worse performance compared to the method proposed here, as can be seen in Table~\ref{table:results_msra}. This is again because such methods are designed specifically for the solely detection of English horizontal text.

\setlength{\tabcolsep}{4pt}
\begin{table}[h]
\begin{center}
\caption{Localization results in the ICDAR 2003 dataset.}
\label{table:results_icdar3}
\begin{tabular}{l c c c}
\hline\noalign{\smallskip}
Method & Precision & Recall & F-score\\
\noalign{\smallskip}
\hline
\noalign{\smallskip}
 \cellcolor{blue!25}\textbf{Ours} & \cellcolor{blue!25}0.74 & \cellcolor{blue!25}0.65 & \cellcolor{blue!25}0.69\\
TD-Mixture \cite{yao2012detecting} & 0.69 & 0.66 & 0.67\\
TD-ICDAR  \cite{yao2012detecting} & 0.68 & 0.66 & 0.66\\
Epshtein \etal~\cite{Epshtein2010}& 0.73 & 0.60 & 0.66\\
Yi \etal  \cite{yi2011}&0.71 & 0.62 & 0.62\\
Becker \etal  \cite{Lucas2003}&0.62 & 0.67 & 0.62\\
Chen \etal~\cite{Chen2004}& 0.60 & 0.60 & 0.58\\
Zhu \etal  \cite{Lucas2003}&0.33 & 0.40 & 0.33\\
Kim \etal  \cite{Lucas2003}&0.22 & 0.28 & 0.22\\
Ezaki \etal  \cite{Lucas2003}&0.18 & 0.36 & 0.22\\
\hline
\end{tabular}
\end{center}
\end{table}
\setlength{\tabcolsep}{1.4pt}

\section{Conclusions}
This paper presents a scene text extraction method in which the exploitation of the hierarchical structure of text plays an integral part. We have shown that the algorithm can efficiently detect text groups whith arbitrary orientation in a single clustering process that involves: a learned optimal clustering feature space for text region grouping, novel discriminative and probabilistic stopping rules, and a new set of features for text group classification that can be efficiently calculated in an incremental way.

Experimental results demonstrate that the presented algorithm outperforms other state of the art methods in three multi-script and arbitrary oriented scene text standard datasets while it stays competitive in the more restricted scenario of horizontally-aligned English text ICDAR dataset. 
Moreover, the presented results in all datasets are obtained with a single (mixed) training set, demonstrating the general purpose character of the method which yields robust performance in a variety of distictly different scenarios.

Finally, the baseline analysis of the algorithm reveals that overall system recall can be substantially increased if needed by using feature space diversification.

\section*{Acknowledgment}
\label{sec:acknowledgement}
This project was supported by the Spanish project TIN2011-24631 the fellowship RYC-2009-05031, and the Catalan government scholarship 2013FI1126.

\ifCLASSOPTIONcaptionsoff
  \newpage
\fi



%
%
%
\bibliographystyle{IEEEtran}
\bibliography{IEEEabrv,tip2014submission}

\newpage

%

\begin{IEEEbiography}[{\includegraphics[width=1in,height=1.25in,clip,keepaspectratio]{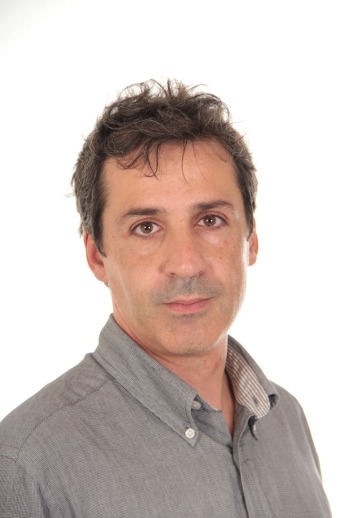}}]{Llu\'is G\'omez i Bigorda\`a}
received the B.S. and M.S. degrees in Computer Science from Universitat Oberta de Catalunya. He received his M.S. degree in Computer Vision and Artificial Intelligence in 2010 from Universitat Autònoma de Barcelona, where he is currently a PhD Candidate and a research assistant in the Computer Vision Center within the Document Analysis and Pattern Recognition Group. 

His research interests are on computer vision and scene text detection and recognition.
\end{IEEEbiography}

\begin{IEEEbiography}[{\includegraphics[width=1in,height=1.25in,clip,keepaspectratio]{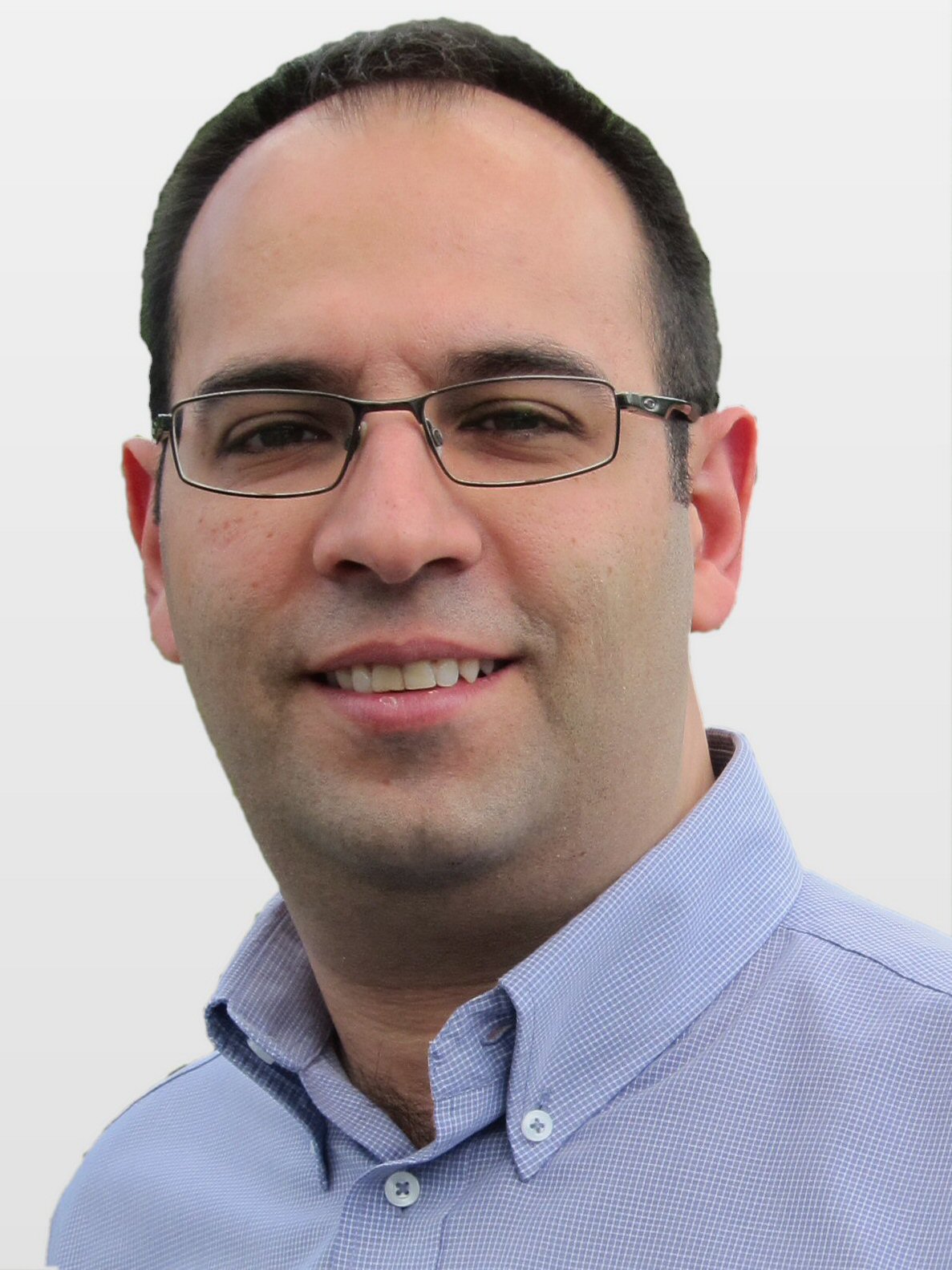}}]{Dimosthenis Karatzas}
Dimosthenis Karatzas is a Senior Research Fellow at the Computer Vision Centre, Universitat Autónoma de Barcelona, Spain. He received his PhD from the University of Liverpool, UK in 2003.
 
He has over 70 publications in the areas of computer vision, document image analysis and colour science. He has considerable experience in robyst reading systems and has been the principal investigator of a number of related research and knowledge transfer projects. In 2013, he received the IAPR / ICDAR Young Investigator Award.
 
Dr Karatzas is a director and co-founder of the spin-off company TruColour, which specializes on perception-based color calibration solutions. Dr Karatzas is the vice-chair of IAPR TC-11 (Reading Systems), a member of the IAPR-Industrial Liaison Committee and member of the IEEE, the SPIE and the IAPR.
\end{IEEEbiography}


\vfill


\end{document}